\documentclass[a4paper,conference]{IEEEtran}
\ifCLASSINFOpdf
\else
\fi
\usepackage{mathtools} 
\usepackage{amsmath,amssymb}

\DeclareMathOperator{\EX}{\mathbb{E}}% expected value
\DeclarePairedDelimiter{\norm}{\lVert}{\rVert} 

\usepackage{float}
\usepackage{caption}

\usepackage{eucal}

\usepackage{adjustbox}

\usepackage{lipsum}
\usepackage{multirow}
\usepackage{xcolor}

\usepackage{algorithmic}
\usepackage{array}

\usepackage{stfloats}
\usepackage{url}
\begin{document}
%
% paper title
% Titles are generally capitalized except for words such as a, an, and, as,
% at, but, by, for, in, nor, of, on, or, the, to and up, which are usually
% not capitalized unless they are the first or last word of the title.
% Linebreaks \\ can be used within to get better formatting as desired.
% Do not put math or special symbols in the title.
\title{Deep-Disaster: Unsupervised Disaster Detection and Localization Using Visual Data}

% author names and affiliations
% use a multiple column layout for up to three different
% affiliations
\author{\IEEEauthorblockN{Soroor Shekarizadeh$^{1}$, Razieh Rastgoo$^{1,2}$, Saif Al‑Kuwari$^{3}$, Mohammad Sabokrou$^{1}$}
\IEEEauthorblockA{$^1$Institute for Research in Fundamental Sciences (IPM)~~ $^2$Semnan University~~ \\$^3$ College of Science and Engineering,
Hamad Bin Khalifa University, Qatar Foundation, Doha, Qatar\\
Email: soroor.shekarizade@gmail.com, rrastgoo@semnan.ac.ir, smalkuwari@hbku.edu.qa, sabokro@ipm.ir}}

% conference papers do not typically use \thanks and this command
% is locked out in conference mode. If really needed, such as for
% the acknowledgment of grants, issue a \IEEEoverridecommandlockouts
% after \documentclass

% for over three affiliations, or if they all won't fit within the width
% of the page, use this alternative format:
%
%\author{\IEEEauthorblockN{Michael Shell\IEEEauthorrefmark{1},
%Homer Simpson\IEEEauthorrefmark{2},
%James Kirk\IEEEauthorrefmark{3},
%Montgomery Scott\IEEEauthorrefmark{3} and
%Eldon Tyrell\IEEEauthorrefmark{4}}
%\IEEEauthorblockA{\IEEEauthorrefmark{1}School of Electrical and Computer Engineering\\
%Georgia Institute of Technology,
%Atlanta, Georgia 30332--0250\\ Email: see http://www.michaelshell.org/contact.html}
%\IEEEauthorblockA{\IEEEauthorrefmark{2}Twentieth Century Fox, Springfield, USA\\
%Email: homer@thesimpsons.com}
%\IEEEauthorblockA{\IEEEauthorrefmark{3}Starfleet Academy, San Francisco, California 96678-2391\\
%Telephone: (800) 555--1212, Fax: (888) 555--1212}
%\IEEEauthorblockA{\IEEEauthorrefmark{4}Tyrell Inc., 123 Replicant Street, Los Angeles, California 90210--4321}}

% use for special paper notices
%\IEEEspecialpapernotice{(Invited Paper)}

% make the title area
\maketitle

% As a general rule, do not put math, special symbols or citations
% in the abstract
\begin{abstract}
Social media plays a significant role in sharing essential information, which helps humanitarian organizations in rescue operations during and after disaster incidents. However, developing an efficient method that can provide rapid analysis of social media images in the early hours of disasters is still largely an open problem, mainly due to the lack of suitable datasets and the sheer complexity of this task. In addition, supervised methods can not generalize well to novel disaster incidents. In this paper, inspired by the success of Knowledge Distillation (KD) methods, we propose an unsupervised deep neural network to detect and localize damages in social media images. Our proposed KD architecture is a feature-based distillation approach that comprises a pre-trained teacher and a smaller student network, with both networks having similar GAN architecture containing a generator and a discriminator. The student network is trained to emulate the teacher's behavior on training input samples, which, in turn, contain images that do not include any damaged regions. Therefore, the student network only learns the distribution of \emph{no damage} data and would have different behavior from the teacher network facing damages. To detect damage, we utilize the difference between features generated by two networks using a defined score function that demonstrates the probability of damages occurring. Our experimental results on the benchmark dataset confirm that our approach outperforms state-of-the-art methods in detecting and localizing the damaged areas, especially for novel disaster types\footnote{Source code is available at: https://github.com/soroorsh/deep-disaster}.
\end{abstract}

% no keywords

% For peer review papers, you can put extra information on the cover
% page as needed:
% \ifCLASSOPTIONpeerreview
% \begin{center} \bfseries EDICS Category: 3-BBND \end{center}
% \fi
%
% For peerreview papers, this IEEEtran command inserts a page break and
% creates the second title. It will be ignored for other modes.

\IEEEpeerreviewmaketitle

% \subsection{Subsection Heading Here}
% Subsection text here.

% \subsubsection{Subsubsection Heading Here}
% Subsubsection text here.

\section{Introduction}
% no \IEEEPARstart
% This demo file is intended to serve as a ``starter file''
% for IEEE conference papers produced under \LaTeX\ using
% IEEEtran.cls version 1.8b and later.
% % You must have at least 2 lines in the paragraph with the drop letter
% % (should never be an issue)
% I wish you the best of success.

% \hfill mds

% \hfill August 26, 2015
Natural and human-caused disaster incidents result in considerable damage every year and affect thousands of people. Sadly, loss of life and physical destruction in such catastrophic events are inevitable. In such emergency situations, saving people's life requires rapid rescue operations, which in turn needs a real-time system to quickly process information to examine the disaster and provide clear insights on what decisions should be made.

In recent years, breakthrough results in deep learning, especially in computer vision and natural language processing, alongside the significant growth of social media platforms (e.g., Twitter, Instagram) provide great opportunities for developing fast and accurate networks by collecting and analysing disaster data, which will help humanitarian organizations in rescue operations. 
% In particular, published content such as texts, images, or videos on social media during disasters would be a rich source to assess the extent of the catastrophe, send support, and identify urgent needs. %Another consideration is the unexpectedness of natural disasters, which is an inherent feature and causes injuries and shock. %
% In urgent conditions, saving people's life requires rapid rescue operations and brings up the need for a real-time system to quickly process the information to examine the disaster and provide clear insights about what decisions should be made.
% In recent years, breakthrough results in deep learning, especially in computer vision and natural language processing, provide opportunities for us to pay our debt to human societies by developing fast and accurate networks to help humanitarian organizations in rescue operations. Alongside, the significant growth of social media platforms (e.g., Twitter, Instagram) facilitated suitable conditions for researchers to collect valuable benchmarks to enable an analysis of the disaster data. 
However, while some works highlighted that the social media imagery data are very informative and can help humanitarian organizations during disasters, there is less focus on the visual data. This is due to the complexity of information extraction from visual data compared to text data (e.g., tweets from Twitter) \cite{DBLP:journals/corr/abs-2011-08916}.

Currently, there are four publicly available datasets on natural disasters: Damage Severity Assessment Dataset (DAD) \cite{9069136}, CrisisMMD \cite{DBLP:journals/corr/abs-1805-00713}, Multi-modal Damage Identification Dataset \cite{mouzannar2018damage}, and MEDIC \cite{alam2021medic}. These datasets have annotated labels for different classification tasks, such as disaster type detection, whether an image is informative or not, categorizing humanitarian aids, and damage severity assessment \cite{DBLP:journals/corr/abs-2011-08916}.  As their data is labeled, these datasets are mainly used to train supervised learning algorithms \cite{9069136, DBLP:journals/corr/abs-2011-08916}.
Nevertheless, a shortcoming of classification approaches is the \emph{cold start} problem, which refers to the necessity of requiring annotated data for the damage assessment task, and  such methods are not able perfectly detect the unseen (novel) damages. In addition, real-time labeling data is pricey and even impractical in some situation. Therefore, these problems make classification models impractical for timely response  \cite{Zhang2020AHT, 9069136}.

One solution to address these problems is domain adaptation approaches \cite{Zhang2020AHT, li2019identifying}. In these approaches, the aim is to classify unlabeled target data by learning from annotated source disaster events that occurred earlier. These methods solve the problem of requiring real-time annotation of train data, but they need a large amount of labeled source data (approximately as much as unlabeled target data for training). Furthermore, transforming the domain during the training process leads to the loss of domain-specific information; this transformation drastically affects the classifier's performance on the target domain. Besides all these approaches, only two studies focus on localizing damages in the disaster images  \cite{DBLP:journals/corr/abs-1806-07378,Li2019LocalizingAQ}. 

To address the mentioned challenges, we propose  \textbf{Deep-Disaster}, an unsupervised method based on deep learning models. Here, we aim to train an unsupervised network to detect damaged areas on unlabeled input images and obtain fast and accurate results, which is necessary in times of emergency. Our problem is similar to a typical anomaly detection problem, where we need to identify abnormality (i.e. disasters) by examining unclassified traffic. Inspired by \cite{DBLP:journals/corr/abs-2011-11108}, Deep-Disaster uses the Knowledge Distillation (KD) method to distill the comprehensive knowledge of a pre-trained teacher network into a smaller student network and does not need any labeled data for training. A pre-trained teacher network is widely used in the KD methods, called offline distillation methods. The main advantage of using a pre-trained network is utilizing the power of teacher's feature representation, especially for the datasets where the size of normal data is small while containing a large variety of different samples \cite{DBLP:journals/corr/abs-2011-11108, wang2021knowledge}.

The student network learns the manifold of the train data comprehensively by forcing some of the student's critical layers to mimic the teacher's. As a result, the student network only learns the \emph{no damage} data distribution and will behave differently from the teacher when facing images containing damaged areas since it does not know other possible input data.
Furthermore, using a smaller network architecture helps the student learn only discriminative features and avoid learning non-discriminative ones during training, leading to more visible different behavior of both networks on damaged areas.

%In summary, the main contributions of this paper: 
\subsection{Contributions}
The main contributions of this paper can be summarized as follow:
\begin{enumerate}
    \item We propose an unsupervised deep network for detecting and localizing disaster-damaged areas from visual data. Our approach generalizes well to unseen and new types of disasters. To the best of our knowledge, Deep-Disaster is one of the first unsupervised works on social media disaster images.
    \item We utilize the KD approach to train only on \emph{no-damage} data by transferring the knowledge of a pre-trained teacher network to a student network. Consequently, the model focuses only on the distinguishable features between \emph{damage} and \emph{no damage} images.
    \item Our unsupervised approach obtains comparable results to the supervised methods in terms of detection and localization.
\end{enumerate}

%\subsection{Organization}
%The rest of the paper is organized as follows: in \ref{related_work} we briefly discuss related work. We we present our proposed model in section \ref{DD}, followed by our experimental results in section \ref{DD_results}. Finally, we conclude the paper in section \ref{conclusion}.

%%%%%%%%%%%%%%%%%%%%%%%%%%%%%%%%%%%%%%%%%%%%%%%%%%%%%%%%%%%%%%
\section{Related Works}\label{related_work}
In this section, we briefly review existing and related research on identifying disaster damage, as well as Knowledge Distillation (KD).

\subsection{Identifying Disaster Damage}
Related literature seems to be focusing on textual data rather than a visual one. However, many studies report on the importance of images posted on social media during such disaster incidents \cite{Alam2017Image4ActOS,9069136,alam2021medic, DBLP:journals/corr/abs-2011-08916,DBLP:journals/corr/NguyenAOI17}.
%In damage assessment, the goal is to estimate the damage severity on the images data, which would help the rescue organization make an informed decision. 
Most studies classify images into three class labels (little/none, mild, severe), and mainly use transfer learning approaches \cite{Alam2017Image4ActOS,DBLP:journals/corr/NguyenAOI17,9069136}, or report the damage severity as a continuous value \cite{DBLP:journals/corr/abs-1806-07378, nia2017building}. 

In addition, there are unsupervised domain adaptation approaches \cite{Zhang2020AHT, li2019identifying} that examine the damage severity by applying a domain adaption framework. These works determine two different disaster events as the source and target data and aim to accurately identify the damaged areas for a target disaster, while only the source feature representation is considered for the classification task. %For example, \cite{li2019identifying} utilizes the Domain-Adversarial Neural Network (DANN)  \cite{ganin2016domain} with the VGG-19 \cite{simonyan2014very} network to transfer the source and the target feature representations in order to make them indistinguishable. %At the same time, only the source feature representation is considered for the classification task. %This approach classifies images into two categories: \emph{damage} and \emph{no damage}. %\color{blue}In the case of having different types of source and target disasters, this model outperforms the VGG-19 model \cite{simonyan2014very}.
%\color{black}
%Recently, \cite{Zhang2020AHT} provided a hybrid deep transfer learning framework called SocailTrans to assess the damage severity level of the images. SocailTrans consists of two modules, an Adversarial Co-Training Transfer Learning (ATLL) module and a Multitask Joint Network Optimization (MJNO) module, and assigns three labels (Severe, Mild, and None) to the unlabeled target data by learning from an annotated source disaster event that occurred earlier.

%In this model, three labels (Severe, Mild, and None) are assigned to the unlabeled target data by migrating a damage assessment model that was learned from an annotated source disaster event that occurred earlier. %\color{blue} In \cite{DBLP:journals/corr/abs-2011-08916}, the authors provided a consolidated dataset by combining and deduplicating the three existing datasets \cite{9069136,DBLP:journals/corr/abs-1805-00713, mouzannar2018damage}. Data in this dataset are relabeled for four different important tasks for humanitarian aid in disaster events. Recently, \cite{DBLP:journals/corr/abs-1805-00713} completed the \cite{DBLP:journals/corr/abs-2011-08916} work by proposing the MEDIC dataset to address the disaster events in a multitask learning framework. 
\color{black}
Moreover, some studies considered localizing damages in social media images and then calculating damage severity using the detected areas \cite{DBLP:journals/corr/abs-1801-09454,YUAN2018758,10.1111/mice.12263,DBLP:journals/corr/abs-1806-07378,Li2019LocalizingAQ}. Specifically, in \cite{Li2019LocalizingAQ, DBLP:journals/corr/abs-1806-07378}, the authors first classify data into two classes (\emph{no damage} or \emph{damage}) using the VGG-19 model \cite{simonyan2014very}. Then, the damaged areas in an image are localized using a class activation mapping (CAM) approach, namely Grad-CAM \cite{DBLP:journals/corr/SelvarajuDVCPB16} and Grad-CAM++  \cite{8354201}. Furthermore, a continuous damage severity assessment entitled Damage Assessment Value (DAV) has been proposed.
% However, a disadvantage of domain adaptation approaches is that they need a large amount of training data of the target disaster event to have a good model which could properly transfer the damage features of the target event. Furthermore, transforming the domain during the training process leads to the loss of domain-specific information; this transformation crucially affects the classifier’s performance on the target domain. Besides all these approaches

\subsection{Knowledge Distillation (KD)}

The intuition behind KD is that a smaller student network is generally trained under the supervision of a larger teacher network \cite{hinton2015distilling}.  KD can be applied to different fields in Artificial Intelligence (AI), such as computer vision, speech recognition, and natural language processing \cite{gou2021knowledge}. There is a vast amount of studies in this area \cite{gou2021knowledge,wang2021knowledge}; however, we focus only on a particular set of more related methods \cite{DBLP:journals/corr/abs-2011-11108,mishra2021vt,wang2021student,bergmann2020uninformed}. In these approaches, a student network learns to mimic a pre-trained feature extractor teacher network during the training process. After that, it estimates the anomalies using a scoring function. In \cite{DBLP:journals/corr/abs-2011-11108}, the authors defined a KD architecture containing a VGG-16 network \cite{simonyan2014very} as a pre-trained teacher network and a smaller student network. They then proposed a novel loss function to teach the student network using intermediate representations of some critical layers corresponding to only normal data. %Hence, a new direction is learned to detect and localize anomalies.
Similarly, the authors in \cite{wang2021student} utilized a feature-based distillation method to train a student network with the helping of a pre-trained equal-size teacher. In this method, the student network receives multi-level knowledge from the feature pyramid under the teacher's supervision. A scoring function is then defined according to the difference between feature pyramids generated by the two networks for anomalies detection.

%%%%%%%%%%%%%%%%%%%%%%%%%%%%%%%%%%%%%%%%%%%%%%%%%%%%%%%%%%%%%%
\section{Proposed Method}\label{DD}

\begin{figure}[ht]
    \centering
    \includegraphics[width=\linewidth]{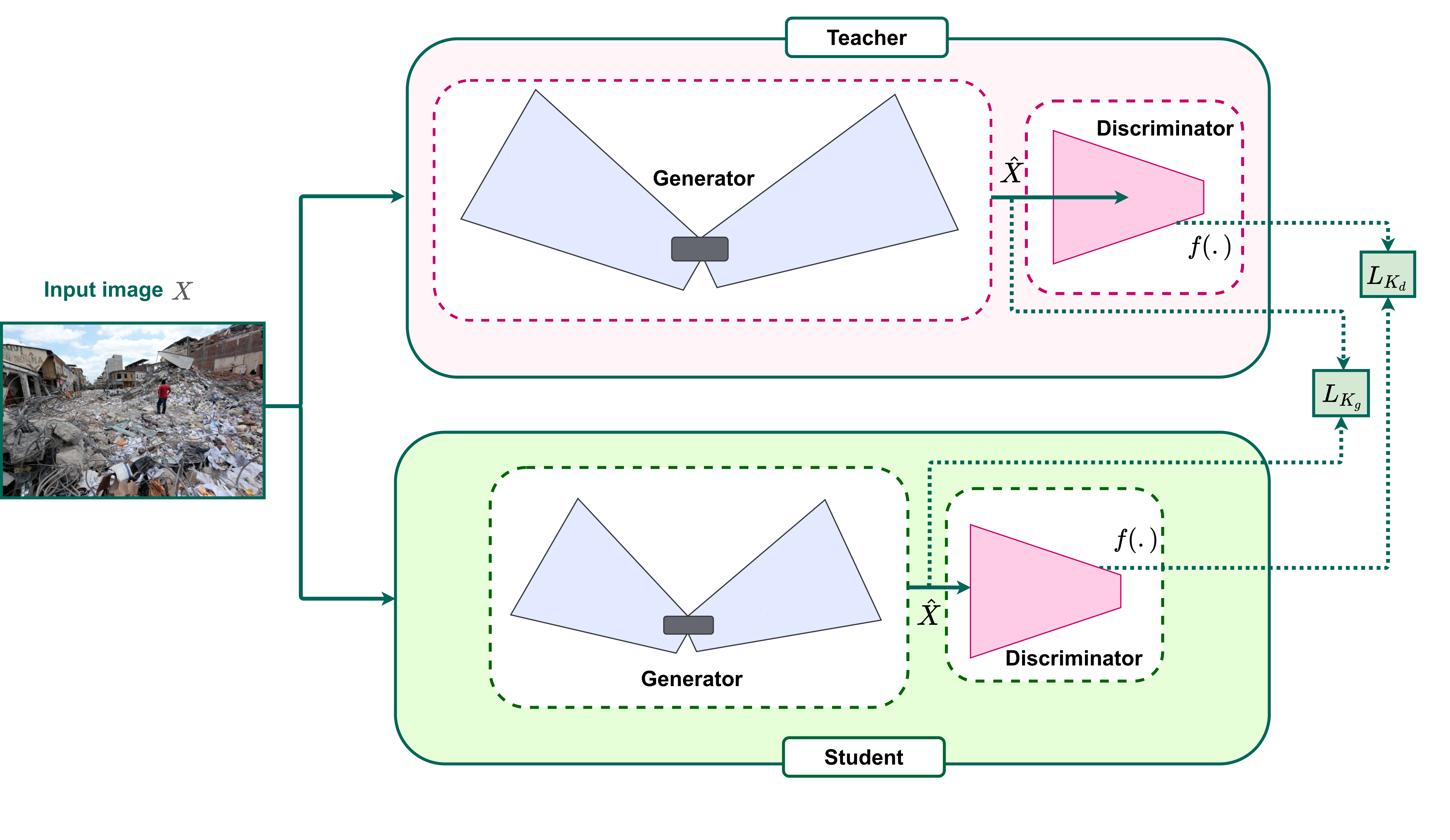}
    \caption{Overview of the proposed method}
    \label{fig:model_arc}
\end{figure}

As previously mentioned, we propose an unsupervised approach for disaster damage detection and localization (i.e, \emph{Deep-Disaster}). Deep-Disaster is a KD framework comprising two networks $S$ and $T$, where $S$ is a student network that learns to replicate the behavior of the pre-trained teacher network $T$ during the training process.
The overall model architecture is depicted in Figure \ref{fig:model_arc}, which was inspired by \cite{DBLP:journals/corr/abs-2011-11108}. Briefly, in our distillation framework, each network is a Convolutional Neural Network (CNN) with the same layers. However, the student network has fewer channels in each layer. Each network consists of a generator $G$ and a discriminator $D$ inspired by \cite{DBLP:journals/corr/abs-1901-08954}. The generator $G$ tries to reconstruct input images to fool the discriminator $D$. The discriminator's task is to classify the input sample into either a real image $x$ or a generated image $\Hat{x}$. %\color{blue}Therefore, we train the student network in an unsupervised adversarially approach using a KD framework. 
\color{black} For a given dataset $D_{train}$, which contains only \emph{no damage} images, we aim to train our student network to learn the distribution of the training samples with the help of the pre-trained teacher network in the KD framework. Then, we evaluate the proposed model on the $D_{test}$ data that includes both \emph{no damage} and \emph{damage} images.
 
\subsection{Student-Teacher Architecture}
\label{ssec:arc}
The basic architecture of the student and teacher networks is similar, which helps in preventing information loss while tackling complex data \cite{wang2021knowledge}. The only difference is in the number of channels in each layers, where the student network has fewer channels in each layer than the teacher's.

\begin{figure}[ht]
    \centering
    \includegraphics[width= 0.9\linewidth]{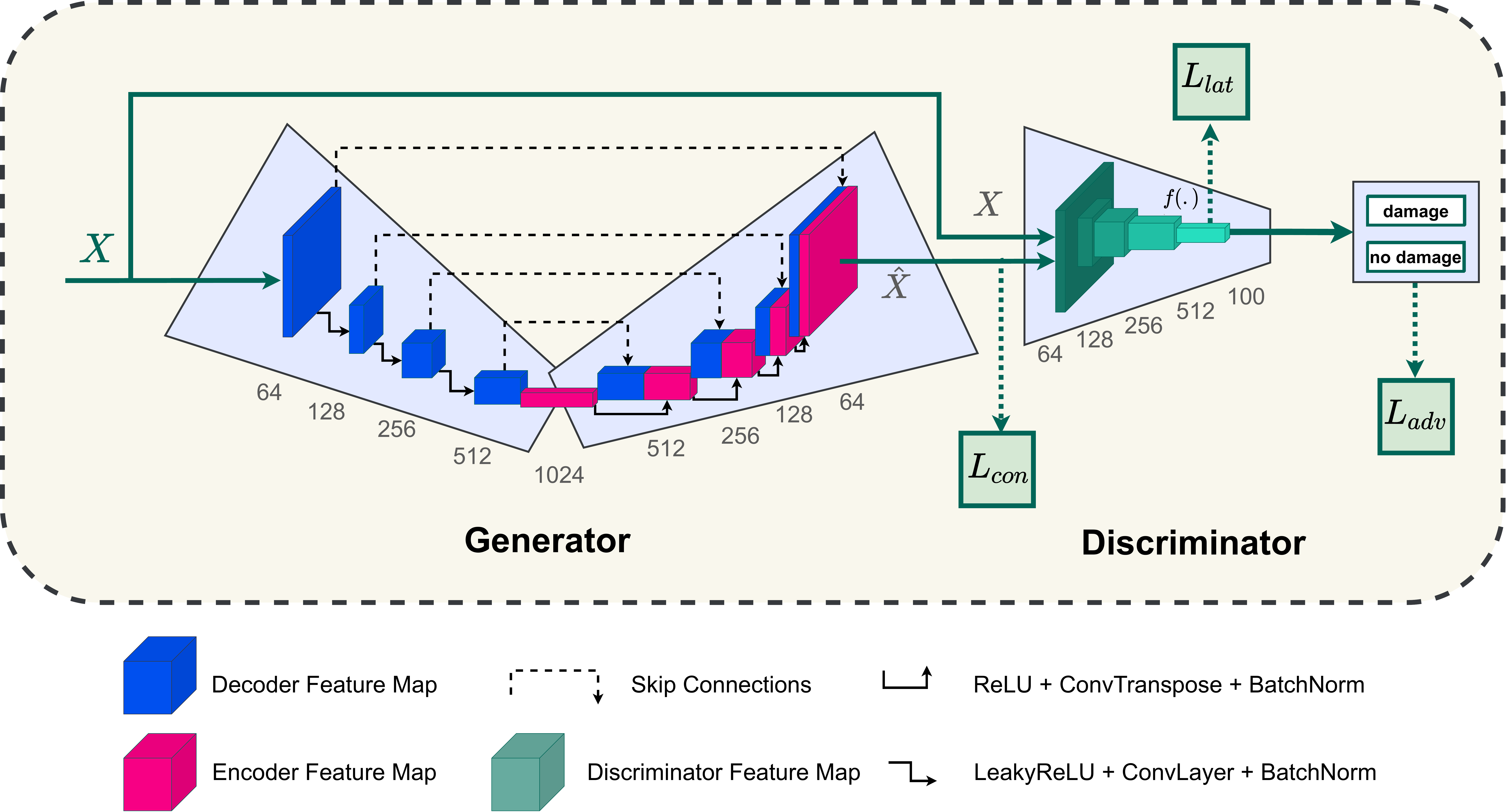}
    \caption{Details of the student network architecture.}
    \label{fig:std_arc}
\end{figure}

As Figure \ref{fig:std_arc} illustrates (inspired by \cite{DBLP:journals/corr/abs-1901-08954}), the student network architecture consists of a U-net \cite{DBLP:journals/corr/RonnebergerFB15} generator $G$  and a discriminator $D$ network.  The generator has an encoder-decoder structure in which the encoder learns the discriminative feature representation of the training data by mapping the high dimensional input image $x$ into a low-dimension feature space $z$. The encoder's role is to reconstruct the input image from the latent representation $z$. However, to have a more precise reconstructed image $\Hat{x}$, the encoder layers are concatenated with their corresponding layers in the decoder network.

Interestingly, using the encoder-decoder structure alongside the skip-connections makes our student network robust in reconstructing complicated input images. %This advantage comes from the skip-connections employed in the model. 
These connections lead to a direct transfer of knowledge between layers while preserving local and global information. Besides the generator network, the discriminator architecture is an encoder adopted from the discriminator structure of DCGAN \cite{Radford2016UnsupervisedRL}. The goal of this network is to correctly distinguish real image $x$ from generated image $\Hat{x}$  through the powerful network ${G}$. %\color{blue}It can be just like the Vanilla GAN \cite{NIPS2014_5ca3e9b1} structure. \color{black} 
In addition, the discriminator obtains latent representations of the input image, regardless of whether it is classified as $x$ or $\Hat{x}$.

\subsection{Training Student-Teacher}
%\color{blue}In contrast to the previous KD methods, where the student network is trained to obtain similar outputs to the teacher, \color{black} 
To apply KD framework, we extend the idea in \cite{DBLP:journals/corr/abs-2011-11108} and employ intermediate feature representations of the teacher network to train the student network. In other words, the student network learns the full capability of the teacher model to generate training samples $\Hat{x}$ and reconstruct latent representations for the input $x$ as close as possible. Motivated by \cite{DBLP:journals/corr/abs-2011-11108}, having a full knowledge transfer from teacher to student relies on considering the value and the direction of the activation functions’ values. 

We assume that two activation vectors with similar Euclidean distances from a source vector do not necessarily have equal outputs in activating the following neuron. As a result, being in the same direction as well as the same Euclidean distance would result in more reliable knowledge transfer. For a selected critical layer $C_m$, we have:

\begin{equation}
  L_{val} = {\frac{1}{N_i}\sum_{j=1}^{N_i}{(a_S^{{C_{l}}_i}(j) -a_T^{{{C}_{l}}_i}(j))^2}}
  \label{eq:val_loss}
\end{equation}

\begin{equation}
  {L}_{dir} = 1-\sum_i{\frac{vec(a_S^{{C_l}_i})^T.vec(a_T^{{C_l}_i})}{\norm*{vec(a_S^{{C_l}_i})}\; \norm*{vec(a_T^{{C_l}_i})}}} 
  \label{eq:dir_loss}
\end{equation}

where $N_i$ denotes the number of the neurons in layer ${C_l}_i$, $a_{.}^{{C_l}_i}(j)$ is the value of $j^{th}$ activation in layer ${C_l}_i$, and the $vec(x)$ indicates a vectorize function.

Therefore, to induce the student network to accurately emulate the teacher’s feature representations on input training samples, we select two critical layers as hints for transferring teacher knowledge. As depicted in Figure \ref{fig:model_arc}, the first one is the latent representations of discriminator ${D}$, which is defined here as ($f(.)$). The second one is the reconstructed image by generator $G$, which is defined as $\Hat{x}$. For each critical layer $C_m$, $L_{val}$ tries to decrease the Euclidean distance, while $L_{dir}$ is used to increase the cosine similarity between the teacher and student output activation values. The KD losses, $K_g$ and $K_d$, are defined in Equation \ref{eq:kg_loss} and Equation \ref{eq:kd_loss}, respectively.

\begin{equation}
  L_{K_g} = L_{val}(\Hat{x}_S, \Hat{x}_T) + \alpha L_{dir}(\Hat{x}_S, \Hat{x}_T)
  \label{eq:kg_loss}
\end{equation}

\begin{equation}
  L_{K_d} = L_{val}(f_S(\Hat{x}),f_T(\Hat{x})) + \alpha L_{dir}(f_S(\Hat{x}),f_T(\Hat{x}))
  \label{eq:kd_loss}
\end{equation}

 where %\color{blue} $f(.)$ is the latent representations of discriminator $D$, $\Hat{x}$ is the reconstructed image by generator $G$, and  \color{black} 
 $\alpha$ is used to have the same range in the loss values. To choose the optimal value of $\alpha$, we consider the initial error amount before training for both terms \cite{DBLP:journals/corr/abs-2011-11108}. 

Moreover, these critical layers are crucial for the student's adversarial training. 
Inspired by \cite{DBLP:journals/corr/abs-1901-08954}, we use a combination of three losses ($L_{adv}$, $L_{con}$, $L_{lat}$) as the adversarial training objective. The Adversarial loss is the well-known MiniMax GAN loss, which helps the model generate realistic images. Contextual loss denotes the reconstruction error of the generated image $\Hat{x}$. The Latent loss is defined to ensure that the network can contextually learn good latent representations of input data. 

\begin{equation}
  L_{adv} = {\EX}_{x {\sim} p_x}{[log D(x)]} + {\EX}_{x {\sim} p_x}{[log (1 - D(\Hat{x})]}
  \label{eq:adv_loss}
\end{equation}

\begin{equation}
  L_{con} = \EX_{x \sim p_x}{\norm*{x-\Hat{x}}_1}
  \label{eq:rec_loss}
\end{equation}

\begin{equation}
  L_{lat} = \EX_{x \sim p_x}{\norm*{f(x)-f(\Hat{x})}_2}
  \label{eq:lat_loss}
\end{equation}
%\color{blue} where $f(.)$ is the latent representations of discriminator $D$, and $\Hat{x}$ is the reconstructed image by Generator $D$. 
\color{black}

Consequently, our training objective for the student network is defined as the weighted sum of all the above losses:

\begin{equation}
  L = \lambda_{adv}L_{adv}+ \lambda_{con}L_{con} + \lambda_{lat}L_{lat} + \lambda_{K_g}L_{K_g} + \lambda_{K_d}L_{K_d}
  \label{eq:train_obj}
\end{equation}

where $\lambda_{adv}$, $\lambda_{con}$ and $\lambda_{lat}$
are the parameters that specify the effectiveness of each loss in the total loss.

\subsection{Damage Detection}
\label{ssec: damage_detect}
To detect damage,  we pass the test samples to both the student and the teacher, which allows us to take advantage of the KD method. In particular, since the student network only learned manifold of the training data (\emph{no damage} and has never seen such damaged images), it would represent different behavior on \emph{damage} samples compared to the teacher, and it is likely to fail in reconstructing damages.  Hence, the generator's reconstructed image and the discriminator's latent vector will help to distinguish \emph{damage} data from \emph{no damage} ones because the similarity of outputs between the student and teacher would decrease in these cases. For a given test sample $\dot{x}$, we formulate score $\mathcal{S}$ in Equation \ref{eq:eq_score}:

\begin{equation}
  \mathcal{S}(\dot{x}) = \omega_{L}\mathcal{L}(\dot{x}) + \omega_{R}\mathcal{R}(\dot{x})  + \omega_{VD}(\mathcal{V}(\dot{x})  + %\omega_{D}
  \alpha \mathcal{D}(\dot{x}))
  \label{eq:eq_score}
\end{equation}

where $\mathcal{L}(\dot{x})$ calculates the similarity score between input image $x$ and the student's generator reconstructed image $\Hat{x}$ based on Equation \ref{eq:rec_loss},  $\mathcal{R}(\dot{x})$ calculates the difference between the student's discriminator latent representation of the input image $f(x)$ and the generated image $f(\Hat{x})$ based on Equation \ref{eq:lat_loss}, $\mathcal{V}(\dot{x})$ and $\mathcal{D}(\dot{x})$ measure similarity of teacher's and student's outputs in value and direction on the generated image, according to Equation \ref{eq:val_loss} and Equation \ref{eq:dir_loss}. $\omega_{L}$, $\omega_{R}$, and $\omega_{VD}$ adjust the effectiveness of the score functions in the final score, and $\alpha$ is defined as Equation \ref{eq:kd_loss} and Equation \ref{eq:kg_loss}.

\subsection{Damage Localization}
\begin{figure*}[ht]
    \centering
    \includegraphics[width=0.6\linewidth]{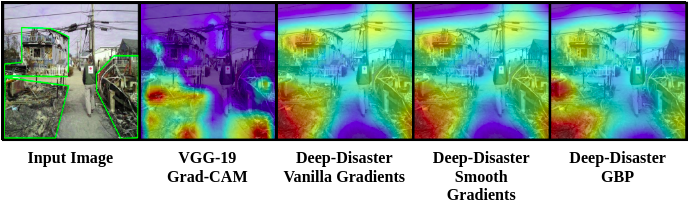}
    \caption{Different localization methods: First column: The input image, Second column: The localization of \cite{DBLP:journals/corr/abs-1806-07378} method, Third-Fifth columns: Our localization results using three different localization methods.}
    \label{fig:local_fig}
\end{figure*}

For the localization task, we investigate three gradient-based interpretability methods, namely: Vanilla Gradient \cite{simonyan2013deep}, Smooth Gradient \cite{DBLP:journals/corr/SmilkovTKVW17} and Guided Back-propagation \cite{springenberg2014striving}. These methods are based on the derivative of loss function w.r.t. the input to highlight the most important pixels. Figure \ref{fig:local_fig} illustrates the results of these methods on \emph{damage} test samples.

\subsubsection{Vanilla Gradient} 
This method first computes a saliency map corresponding to the gradient of an output neuron with respect to the input, highlighting the areas of the given image \cite{linardatos2021explainable}. We apply the Gradient method \cite{simonyan2013deep} on our final loss (Equation \ref{eq:train_obj}) to highlight the damaged areas in the test image samples. We are able to identify damages using this approach because these highlighted regions (i.e., damages) cause an increase in the gradient’s value. For a given input image $x$, the localization map is calculated as follows: 
     
\begin{equation}
      {loc}_{map}(x) =\frac{\partial L}{\partial x} 
  \label{eq:vanilla_grad}
\end{equation}
where $L$ is the training objective function (Equation \ref{eq:train_obj}). 

% Moreover, we apply Gaussian blur and opening morphological filter to reduce the noises in the output localization maps.

\subsubsection{Smooth Gradient}
This method provides a less noisy saliency map by adding several input images perturbed with random Gaussian noise and averaging the resulting noisy gradients. This method extends any gradient-based interpretability method by adding a further step to the approach. 
We apply Smooth Gradient on the Vanilla Gradient Method to obtain a less noisy localization map. This approach is calculated as follows:

\begin{equation}
      {\Hat{Loc}}_{map}(x) = \frac{1}{n} \sum_{1}^{n}{Loc}_{map}(x + \mathcal{N}(0,\,\sigma^{2}))
  \label{eq:smooth_grad}
\end{equation}

\subsubsection{Guided Back-propagation}
This technique only uses the positive gradient with respect to input by back-propagating through the Relu activation function.
As a result, we have only the positive gradients after changing the negative gradients’ values to zero, which means that the input features that activate the neurons would be highlighted.

%%%%%%%%%%%%%%%%%%%%%%%%%%%%%%%%%%%%%%%%%%%%%%%%%%%%%%%%%%%%%%
\section{Experimental Results}\label{DD_results}
We evaluate the performance of Deep-Disaster on the Damage Assessment Dataset (DAD)  \cite{9069136} for detecting and localizing damages in different disaster types. Our model is implemented using PyTorch \cite{NEURIPS2019_9015}. All reported results are performed using an NVIDIA GeForce RTX 2060 GPU.

The initial Learning rate of the student network is equal to $2e^{-3}$ with a decay rate. Also, momentum are equal to 0.5, with a batch size of 64. The hyper-parameters of our training objective function are: $\mathcal{\lambda}_{adv}=1$, $\mathcal{\lambda}_{lat}=1$,  $\mathcal{\lambda}_{cons}=20$, $\mathcal{\lambda}_{K_g}= 50$, and $\mathcal{\lambda}_{K_d}=1$. 
Moreover, we obtain optimum weights of Equation \ref{eq:eq_score} by achieving the highest AUC-ROC using $\omega_{L}=0.4$, $\omega_{R}=0.2$, and $\omega_{VD}=0.4$.

\subsection{Dataset}
As the name implies, the DAD dataset is an imagery dataset to assess the level of damage in disaster events. The DAD contains social media images collected from AIDR \cite{imran2014aidr}. % and Google.
 The images from AIDR are gathered during four different natural disaster events: Philippines Ruby Typhoon (2014), Nepal Earthquake (2015), Ecuador Earthquake (2016), and USA Hurricane Matthew (2016).
%, while the images from Google are crawled using queries like damage building, damage bridge, damage road. 
The dataset contains  $\sim$25K images labeled into three classes of severity levels: severe, mild, and little-to-no damage.

However, following \cite{li2019identifying}, we combine both the mild and severe classes into one class named \emph{damage}, since we aim to train an unsupervised end-to-end network to distinguish \emph{no damage} images from the \emph{damage} ones. The original data splitting is: training (80\%) and test (20\%) \cite{9069136}. 
Since we only used \emph{no-damage} data during training,  we followed the original dataset splitting for \emph{no-damage} data (80\%). Also, for evaluation, we kept the ratio (20\%) for images data with \emph{no-damage} and \emph{damage} labels. Table \ref{tab:dad_split} shows the class distribution for each disaster in the Train and Test phases. 

\begin{table}
  \caption{DAD distribution on each disaster class for the train and test}
  \label{tab:dad_split}
  \centering
  \begin{tabular}{|c|c|c|c|}
    \hline
     \textbf{Disaster name} & \textbf{Class labels} & \textbf{Train} & \textbf{Test}\\
   \hline
    Ruby Typhoon & no damage & 320 & 80  \\
    (486)  & damage & - & 86 \\ 
    \hline
    
    Nepal Earthquake & no damage & 6336 & 1584  \\
    (10156) & damage & - & 2236  \\ 
    \hline
    
    Ecuador Earthquake   & no damage & 730 & 182  \\
    (1186) & damage & - & 274  \\ 
    \hline
    
    Matthew Hurricane & no damage & 261 & 65  \\
    (380)  & damage & - & 54  \\
    \hline
    
  \end{tabular}
\end{table}

\subsection{Evaluation Metric}
The evaluation of our method is measured by the popular AUC-ROC metric. AUC-ROC measures performance for all possible classification thresholds calculating the area under the ROC curve. This metric assesses the capability of a model in distinguishing between classes. As a model approaches AUC-ROC of 1, the model's accuracy improves. We utilize this metric to evaluate how our model performs in discriminating between \emph{no-damage} and \emph{damage} classes. 

\subsection{Ablation Studies}

We conduct a series of ablation studies on the DAD dataset to answer the following questions: Is our Knowledge distillation structure really effective? Is student architecture improve the results (distillation effect)? Which layer/layers are effective while engaging in the training procedure?

\subsubsection{Training Structure}
\begin{figure}[ht]
    \centering
    \includegraphics[width=0.9\linewidth]{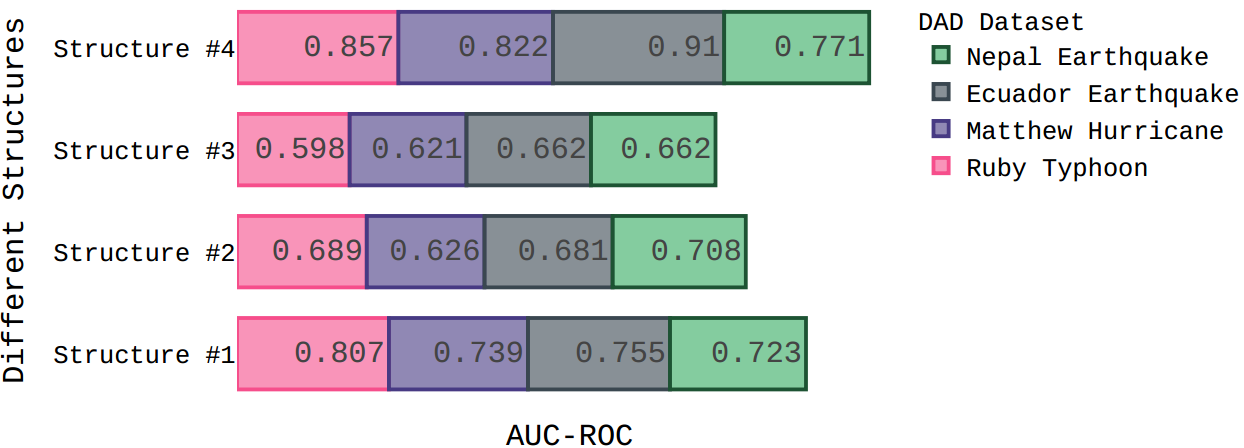}
    \caption{Comparison of AUC-ROC score on all classes for different training structures. % The results show that using a pre-trained teacher for training the student network in the KD frameworks outperforms all the other structures.
    }
    \label{fig:train_diff}
\end{figure}

For the first experiment, we compare four different architectures and training procedures: 
(1) Teacher network only, 
(2) Student network only, 
(3) Teacher and Student simultaneously from scratch, and  (4) Student in the KD method using a pre-trained teacher. Figure \ref{fig:train_diff} shows the AUC-ROC of all classes for these experiments. The results demonstrate that not only could we train a more compact network on the dataset, but also our framework outperforms all the other structures in this experiment. 

\subsubsection{Distillation effect}
For the next experiment, we investigate the effect of the student structure in our KD architecture. In the KD definition, a smaller student than the teacher helps to learn only distinguishing features and obliterating the non-distinguishing ones, especially in our unsupervised manner when there are only \emph{no damage} data during training \cite{DBLP:journals/corr/abs-2011-11108}. 
Table \ref{tab:distill_tab} shows that the smaller student achieves better performance than the equal size student for all classes, as we expected. The results originate from the fact that the images in the DAD dataset are hard to discriminate \cite{DBLP:journals/corr/NguyenAOI17}. 

\begin{table}
  \caption{Comparison of AUC-ROC score on all classes for different student network size.}
  \label{tab:distill_tab}
  \centering
  \begin{adjustbox}{width=\linewidth}
%   \tiny
  \begin{tabular}{|c|cccc|}
    \hline
      & Ruby Typhoon & Matthew Hurricane  & Ecuador Earthquake & Nepal Earthquake \\
    \hline
    \textbf{Smaller} & 0.857& 0.822& 
    0.91 & 0.771 \\
  \hline
    \textbf{Equal} & 0.715 & 0.631 &  0.82 &  0.635 \\
    \hline
  \end{tabular}
  \end{adjustbox}

\end{table}

\subsubsection{Intermediate Knowledge}
\begin{figure}[ht]
    \centering
    \includegraphics[width=\linewidth]{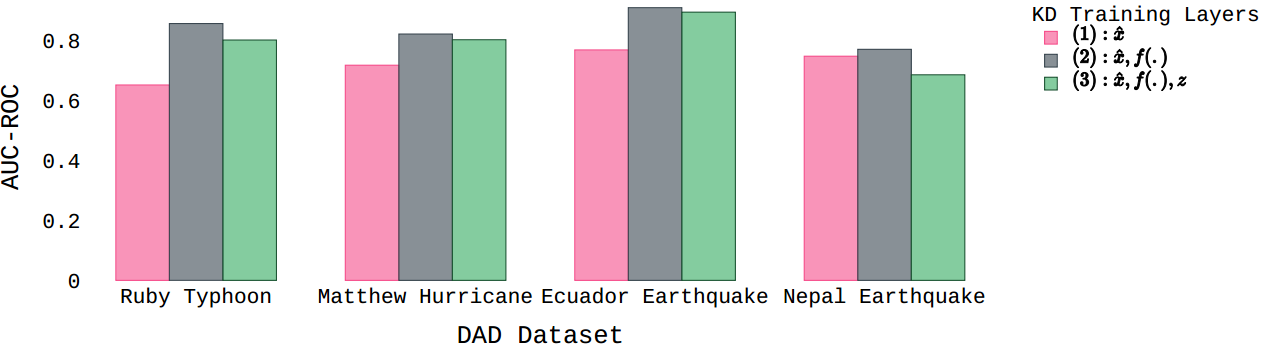}
    \caption{Comparing engaging different layers in the training procedure.% where $\Hat{x}$, $f(.)$, and  $z$ are the generated image, discriminator's last convolutional layer, and  generator’s latent vector for student network, respectively. 
    }
    \label{fig:train_method}
\end{figure}

To examine our KD method, we chose different layers for engaging in the training procedure; Figure \ref{fig:train_method} %shows the results. 
%(1) $\Hat{x}$, (2) $\Hat{x}$ and $f(.)$, (3) $\Hat{x}$, $z$, and $f(.)$. 
%The results
demonstrates that adding the $f(.)$ layer outperforms using solely the $\Hat{x}$ layer; however, adding $z$ layer to the second training configuration does not enhance the results.  
The results confirm that increasing multiple features may not improve distilling knowledge. \cite{wang2021knowledge}

\subsection{Results and Discussions}
\begin{table*}
  \caption{AUC-ROC results on the DAD dataset \cite{9069136}}
  \label{tab:res_tab}
  \centering
  \begin{adjustbox}{width=1\textwidth}
  \small
  \begin{tabular}{|c|c|c|c|c|c|c|}
    \hline 
    \textbf{Category} &
    \textbf{Method}
        & \textbf{Ruby Typhoon}&  \textbf{Nepal Earthquake} & \textbf{Ecuador Earthquake} & \textbf{Matthew Hurricane} & \textbf{Average}\\
       \hline
    % \hline
    \multirow{4}{*}{Supervised} &\textbf{VGG19} \cite{simonyan2014very} & 0.884	& 0.905 & 	0.8846 &	0.765	& 0.85965 \\
  
    &\textbf{VGG19-unseen} \cite{simonyan2014very} & 0.679	&0.758&	0.822&	0.765&	0.756\\
 
    &\textbf{VGG16} \cite{simonyan2014very} & 0.897 &	0.898	&0.9276	&0.828	& 0.88765\\
   
    &\textbf{VGG16-unseen} \cite{simonyan2014very} & 0.655	&0.795&	0.817&	0.768&	0.75875\\
    \hline
    \multirow{4}{*}{Domain Adaptation} &\textbf{DANN} \cite{li2019identifying} & 0.732	&0.738	& 0.795	& 0.779 &	0.761\\ 
       &\textbf{DANN-unseen} \cite{Zhang2020AHT} & 0.639	&0.73&	0.751	&0.72& 0.7105\\
      &\textbf{SocialTrans} \cite{Zhang2020AHT} & 0.742 &	0.779 & 0.839	&0.806	& 0.7915\\
        &\textbf{SocialTrans-unseen} \cite{Zhang2020AHT} &0.688&	0.745&	0.801&	0.756	&0.7475\\
    \hline
    \multirow{2}{*}{Unsupervised} &\textbf{Deep-Disaster}  & 0.857&	0.771&	0.91&	0.822	&0.84\\
     & \textbf{Deep-Disaster-unseen}  & \textbf{0.765} &	\textbf{0.803} &	\textbf{0.945} &	0.701	& \textbf{0.8035}\\
     \hline
  \end{tabular}
  \end{adjustbox}
\end{table*}

Table \ref{tab:res_tab} presents a comparison between the performance of our proposed method and the state-of-the-art methods on the DAD dataset.  As Table \ref{tab:res_tab} shows, we achieved comparable results with the supervised methods and outperform domain adaptation methods.
To the best of our knowledge, we are the first unsupervised approach for disaster detection trained on this dataset. Thus, to have a fair comparison, for each class, we report the average AUC-ROC score of testing on the rest of the classes' trained models.
%between our method and the state-of-the-art methods
%for each class, we report average test results on the %trained state-of-the-art networks on other classes. 
The results marked by the "unseen" label (i.e., SocialTrans-unseen) in the Table \ref{tab:res_tab} illustrate that our method outperforms other methods in three out of four classes for novel disaster types and outperforms them on the overall average AUC-ROC score. %The results denote the generalization capability of the Deep-Disaster approach.

\subsubsection{Localization}
Our localization results in Figure \ref{fig:local_fig} demonstrate that Deep-Disaster outperforms the model in \cite{DBLP:journals/corr/abs-1806-07378}, even though the latter is a supervised method. It shows that our model has learned the disasters and their discriminative features as well, despite the fact that we train it only on \emph{no damage} images.% which leads to increasing the network's attention to the damages.
 However, we could not apply the Grad-CAM approach to our model as it is a supervised method requiring class-related localization weights. The results of the Vanilla Gradients and Smooth Gradients methods are visually similar, but the Smooth Gradients method exhibits less noisy localization since this method calculates an average over gradients of its noisy inputs. 
The GBP approach has competitive results compared to both methods, although it partially does image reconstructing and is unrelated to the network decision-making process \cite{springenberg2014striving}.

%discussion
\subsubsection{Misclassified Images}
\begin{figure}[ht]
    \centering
    \includegraphics[width= 0.6
    \linewidth]{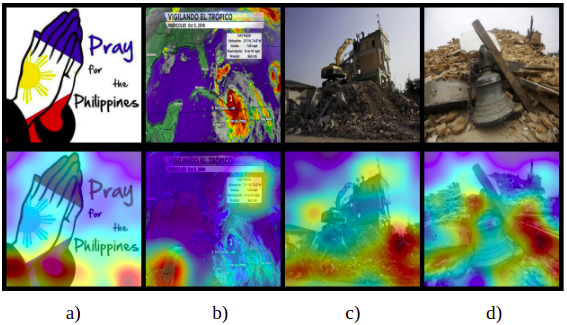}
    \caption{Samples of misclassified images. Each column contains the original image and its localization, respectively}
    \label{fig:miss_fig}
\end{figure}

To investigate images that our model does not recognize, we estimate a threshold on our damage detection score to find which images do not score appropriately \cite{DBLP:journals/corr/abs-2106-16020}. % According to \cite{DBLP:journals/corr/abs-2106-16020}, we define the threshold as a value that gives us an equal number of false positive rate(fpr) and true positive rate(tpr) for both class's labels (\emph{damage} and \emph{no damage}) based on their volumes. 
%\color{blue} So, we could find the answer to this question: Does our framework detect the damaged areas for misclassified images or not? \color{black}
As shown in Figure \ref{fig:miss_fig}(a),(b), our model could not detect some images properly -- most likely this due to lack of training data for some classes %(e.g., ruby typhoon)%
as they contain different kinds of \emph{no damage} images (e.g., images containing text, or satellite images). Additionally, misclassification occurs because our training dataset (Tiny ImageNet) is not well enough to learn all the discriminative features during training the student network.

In addition, some images are not detected since they are very similar to the training samples or contain more than one scene. On the other hand, some images have not scored as images with damaged areas, even thought they were localized correctly (Figure \ref{fig:miss_fig}(c),(d)). These discrepancies usually arise because there is no clear boundary between the definitions of class labels \cite{DBLP:journals/corr/NguyenAOI17}.  %\color{blue} For instance, some images are labeled with the "None or Little" label while they have similar visual features by some images labeled as "Mild." \color{black}
Consequently, in these such cases, the detection scores would not be distinguishable. Another consideration is the presence of identical images, especially in the Nepal Earthquake class, which affects the model's performance drastically. These samples demonstrate the complexity of the DAD dataset. 

%%%%%%%%%%%%%%%%%%%%%%%%%%%%%%%%%%%%%%%%%%%%%%%%%%%%%%%%%%%%%%
\section{Conclusion}\label{conclusion}
In this paper, we propose an efficient unsupervised method for disaster damage detection and localization based on visual data, which we call \emph{Deep-Disaster}. Deep-Disaster is a KD method consisting of a pre-trained teacher and a smaller student network with similar architecture, where the intermediate knowledge of the teacher network is distilled on training images (\emph{no damage} images)
into the compact student network. Then, we detect the damaged areas using the networks' different behavior in values and directions of their critical layers for the test images. % with damaged areas.
In addition, we benefit from interpretability methods to extract localization maps. We used the DAD dataset to evaluate our method. The obtained results confirmed that our proposed method could detect damages without being explicitly trained on such damaged areas and gave an insight into
the generalization capability of our method to novel disaster damages. A possible extension to this work is to conduct more experiments to determine the effectiveness of Deep-Disaster on the other mentioned disaster damage datasets and their various related tasks.

% trigger a \newpage just before the given reference
\newpage
% number - used to balance the columns on the last page
% adjust value as needed - may need to be readjusted if
% the document is modified later
%\IEEEtriggeratref{8}
% The "triggered" command can be changed if desired:
%\IEEEtriggercmd{\enlargethispage{-5in}}

% references section

% can use a bibliography generated by BibTeX as a .bbl file
% BibTeX documentation can be easily obtained at:
% http://mirror.ctan.org/biblio/bibtex/contrib/doc/
% The IEEEtran BibTeX style support page is at:
% http://www.michaelshell.org/tex/ieeetran/bibtex/
%\bibliographystyle{IEEEtran}
% argument is your BibTeX string definitions and bibliography database(s)
% \bibliography{IEEEabrv,../bib/paper}
%
% <OR> manually copy in the resultant .bbl file
% set second argument of \begin to the number of references
% (used to reserve space for the reference number labels box)

% \newpage
% \bibliographystyle{IEEEtran}
% \bibliography{IEEEabrv,ref.bib}
% that's all folks
\end{document}